\documentclass[runningheads]{llncs}

\usepackage[mobile]{eccv}
\usepackage{eccvabbrv}
\usepackage{graphicx}
\usepackage{booktabs}
\usepackage{amsmath}
\usepackage{amssymb}
\usepackage{bm}
\usepackage[accsupp]{axessibility}
\usepackage[ruled,linesnumbered]{algorithm2e}
\usepackage{xcolor}
\usepackage{multirow}
\usepackage{colortbl}
\usepackage{pifont}

\usepackage{pgfplots}
\pgfplotsset{compat=1.18}
\usepackage{subcaption}
\usetikzlibrary{arrows.meta, shapes.geometric, calc, decorations.pathreplacing}

\definecolor{clrInput}{HTML}{4285F4}
\definecolor{clrDpp}{HTML}{137333}
\definecolor{clrSpec}{HTML}{E37400}
\definecolor{clrDual}{HTML}{5F6368}
\definecolor{clrFast}{HTML}{1A73E8}
\definecolor{clrSlow}{HTML}{C5221F}
\definecolor{clrOut}{HTML}{7627BB}
\definecolor{frmA}{HTML}{4285F4}
\definecolor{frmB}{HTML}{34A853}
\definecolor{frmC}{HTML}{FBBC04}
\definecolor{frmD}{HTML}{EA4335}
\definecolor{frmE}{HTML}{A142F4}
\definecolor{frmF}{HTML}{24C1E0}
\definecolor{frmG}{HTML}{F538A0}
\definecolor{frmH}{HTML}{5F6368}

\usepackage{hyperref}
\usepackage{orcidlink}

\newcommand{\method}{DART}

\makeatletter
\def\@fnsymbol#1{\ensuremath{\ifcase#1\or\dagger\or{\star\star}\or
   {\star\star\star}\or \ddagger\or \mathchar "278\or \mathchar "27B\or
   \|\or **\or \dagger\dagger\or \ddagger\ddagger \else\@ctrerr\fi}}
\makeatother

\begin{document}
\title{DART: Difficulty-Adaptive Routing for\\Zero-Shot Video Temporal Grounding}

\titlerunning{DART: Difficulty-Adaptive Routing for Zero-Shot VTG}

\author{Zhengbo Zhang\inst{1}\orcidlink{0009-0000-8956-0095} \and
Mark He Huang\inst{1}\orcidlink{0000-0002-9217-4977} \and
Zhigang Tu\inst{2}\thanks{Corresponding author} \and
Ming-Hsuan Yang\inst{3}\orcidlink{0000-0003-4848-2304}}

\authorrunning{Zhengbo Zhang et al.}
\institute{Singapore University of Technology and Design \and
Wuhan University\and
University of California, Merced}

\maketitle

\vspace{1.5mm}
\begin{center}
\includegraphics[width=\textwidth]{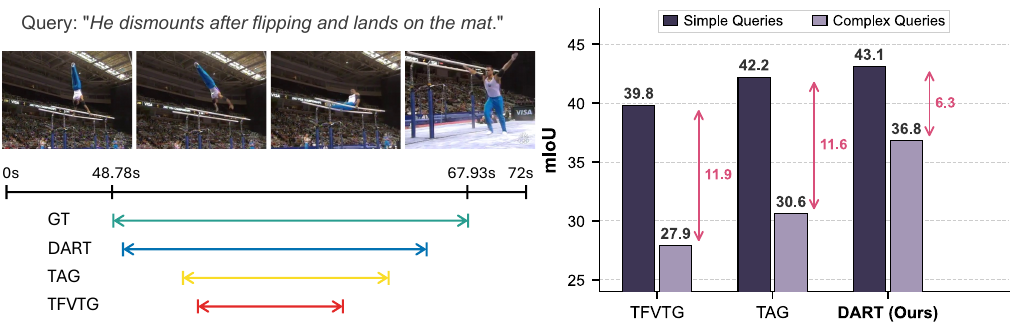}
\captionof{figure}{\textbf{Reasoning gap in zero-shot VTG.} Left: a qualitative example from ActivityNet Captions~\cite{krishna2017activitynet} whose query requires temporal ordering. Feature-matching methods (TFVTG~\cite{zheng2024tfvtg}, TAG~\cite{lee2025tag}) match only ``land,'' missing the earlier flipping phase, while \method{} localizes the full event. Right: mIoU evaluated on 100 simple and 100 complex queries sampled from the ActivityNet Captions val\_2 split. Feature-matching methods exhibit a large performance gap between simple and complex queries; \method{} reduces this gap by nearly half through difficulty-adaptive reasoning.}
\label{fig:teaser}
\end{center}
\vspace{2mm}

\begin{abstract}
Zero-shot video temporal grounding (VTG) localizes events in untrimmed videos from natural language queries without task-specific training.
Existing methods rely on frame-query feature matching, which suffices for simple events but struggles with complex multi-stage queries that require understanding temporal ordering and causal structure---a disparity we call the \emph{reasoning gap}.
We propose \method{} (\textbf{D}ifficulty-\textbf{A}daptive \textbf{R}outing for \textbf{T}emporal Grounding), which bridges this gap by coupling difficulty-aware routing with structured reasoning in large vision-language models.
A query-conditioned Determinantal Point Process (DPP) serves a dual role: selecting diverse, query-relevant keyframes as temporal evidence, and providing spectral entropy as a difficulty indicator.
Simple queries are routed to a Fast path for direct prediction, while complex queries follow a Slow path with Temporal Markup Prompting, which decomposes localization into global event analysis, per-frame temporal role annotation, and boundary extraction.
On Charades-STA and ActivityNet Captions, \method{} achieves state-of-the-art zero-shot performance across both identically distributed and multiple out-of-distribution settings, improving mIoU by up to \textbf{3.5} points over the strongest baseline while using over $7\times$ fewer frames.
The project homepage is available at \url{https://dart-vtg.github.io/}.
\end{abstract}

\section{Introduction}
\label{sec:intro}

Given a natural language query, localizing the target event in an untrimmed video is central to video retrieval~\cite{gao2017tall,lei2022assistsr}, temporal question answering~\cite{lei2018tvqa,wong2022assistq}, highlight detection~\cite{lei2021detecting}, and aligning generated motion, speech, recognizing actions under challenging illumination~\cite{11250624}, and sound~\cite{cheng2026unisonharmonizingmotionspeech,song2026interactiveavatarrealtimestreamingvideo}.
This task, known as video temporal grounding (VTG)~\cite{gao2017tall,krishna2017activitynet,chen2018temporally}, has been extensively studied under fully supervised settings~\cite{gao2017tall,krishna2017activitynet}.
However, supervised methods rely on frame-level temporal annotations, which are both expensive to collect at scale~\cite{mithun2019weakly,tu2025informative} and inherently biased toward the training distribution~\cite{yuan2021closer}.
Zero-shot VTG has therefore attracted growing interest.
Leveraging pre-trained Vision-Language Models (VLMs) and Large Vision-Language Models (LVLMs), recent methods~\cite{luo2024zeroshot,wattasseril2023zeroshot,xu2024vtggpt,zheng2024tfvtg,lee2025tag,jeon2025dbcon,jeon2026granalign} score frame-query similarity via feature matching and aggregate high-scoring frames into temporal proposals, localizing events without any task-specific training data.

Although feature matching has shown promising results on simple queries, it scores each frame independently and cannot capture cross-frame dependencies such as temporal ordering, causal relationships, and event composition~\cite{li2022compositional,yang2023deco,zheng2024tfvtg}.
Complex queries whose localization demands such reasoning therefore remain a significant challenge.

To quantify this limitation, we sample 100 simple queries and 100 complex queries (the latter requiring temporal ordering, causal relations, or conditional logic) from the ActivityNet Captions~\cite{krishna2017activitynet} val\_2 split and evaluate state-of-the-art zero-shot methods on each group.
As shown in \cref{fig:teaser}, the best-performing method drops \textbf{11.6} mIoU points on the complex group (42.2\% \emph{vs.}\ 30.6\%).
For instance, localizing \emph{``He dismounts after flipping and lands on the mat''} requires understanding the temporal ordering of flipping, dismounting, and landing---a capability that per-frame matching lacks.
We term this systematic performance gap the \textbf{reasoning gap}.

A natural way to bridge this gap is Chain-of-Thought (CoT) prompting~\cite{wei2022cot}, which activates the reasoning capabilities that LVLMs already possess~\cite{wang2025timezero} by decomposing complex predictions into explicit intermediate steps, without any task-specific training.
However, applying CoT to temporal grounding is far from straightforward, introducing two key challenges.
\textbf{Challenge~I: How to obtain compact yet sufficient temporal evidence from raw, noisy videos for CoT reasoning?}
Existing zero-shot methods typically feed uniformly sampled or fixed-stride frames to the model~\cite{luo2024zeroshot,zheng2024tfvtg}, yet such query-agnostic sampling suffers from two failure modes.
First, it may miss key event stages, forcing the model to reason over incomplete temporal context~\cite{hu2025mllm}.
Second, it retains many irrelevant frames that act as distractors, overwhelming intermediate reasoning steps and causing errors to compound along the chain~\cite{wei2022cot}.
\textbf{Challenge~II: How to adaptively decide when reasoning is needed?}
Not all queries benefit from multi-step reasoning. A simple query such as \emph{``person sits down''} can be localized by visual matching alone; applying reasoning indiscriminately not only incurs unnecessary latency but also risks compounding errors across intermediate steps, ultimately degrading localization accuracy~\cite{wei2022cot,liu2025mind}.

We propose \textbf{\method{}} (\textbf{D}ifficulty-\textbf{A}daptive \textbf{R}outing for \textbf{T}emporal Grounding), which addresses both challenges within a single framework.
The key observation is that selecting diverse, query-relevant keyframes as temporal evidence naturally exposes how many distinct stages the event contains, which in turn indicates query difficulty---thereby unifying evidence extraction and difficulty routing.

For Challenge~I, we cast temporal evidence extraction as a keyframe selection problem: the selected frames should be both relevant to the query and mutually diverse, covering all stages of the target event while reducing redundancy.
Jointly optimizing for relevance and diversity is non-trivial, because diversity couples every frame's marginal value to the frames already chosen, requiring in principle an exponential search over all subsets.
Inspired by determinantal point processes (DPP)~\cite{kulesza2012dpp}, which provide efficient greedy approximations with provable near-optimality guarantees for diverse subset selection, we design a query-conditioned variant that selects keyframes jointly maximizing query relevance and inter-frame diversity.

Our query-conditioned DPP also addresses Challenge~II.
Complex events that require reasoning comprise multiple sub-events with rich relational structure (\eg temporal ordering, causal dependencies).
Since the DPP kernel encodes this relational structure among all frames, its spectral entropy measures the effective number of dominant components, which in our VTG setting correlates with the complexity of the query-relevant temporal structure.
We use spectral entropy as a difficulty indicator to route each query: queries with low spectral entropy are sent to a \textbf{Fast path} that directly predicts timestamps, while queries with high spectral entropy follow a \textbf{Slow path} that invokes our Temporal Markup Prompting, a structured CoT procedure decomposing grounding into event analysis, per-frame role tagging, and boundary extraction.

To validate the generalization and robustness of \method{}, we evaluate on Charades-STA~\cite{gao2017tall} and ActivityNet Captions~\cite{krishna2017activitynet} under the standard IID setting, multiple OOD settings that vary temporal location bias, dataset split, and query vocabulary, and a cross-dataset transfer setting.
\method{} sets a new state of the art across all settings, with gains of up to \textbf{3.4} mIoU on OOD and compositional splits, confirming that structured reasoning generalizes better than feature matching to novel distributions and queries.

\section{Related Work}
\label{sec:related}

\subsection{Training-based Video Temporal Grounding}

Training-based Video Temporal Grounding methods can be broadly classified into three categories: fully supervised~\cite{gao2017tall,chen2018temporally,zhang2020span,lei2021detecting}, weakly supervised~\cite{mithun2019weakly,gao2019wslln}, and unsupervised approaches~\cite{nam2021zero}.
Fully supervised VTG methods train models using pairs of natural language queries and their corresponding video segments with precisely annotated start and end boundaries~\cite{zhang2020learning,yuan2019semantic}.
While such fine-grained supervision enables accurate localization, the resulting models often inherit biases from the limited and domain-specific nature of manually curated datasets, which restricts their generalization to unseen videos or domains~\cite{yuan2021closer,bao2022learning}.
To mitigate this dependency, weakly supervised and unsupervised paradigms have been introduced~\cite{mithun2019weakly,gao2019wslln,nam2021zero}.
Weakly supervised methods utilize coarse video-level textual descriptions without requiring frame-level temporal annotations~\cite{duan2018weakly,huang2023weakly}, whereas unsupervised approaches eliminate textual supervision entirely by generating pseudo-text queries from the visual content itself and treating them as weak labels~\cite{zheng2023SPL,kim2023language}.
Although these approaches alleviate reliance on exhaustive human annotations, their performance remains constrained by biases in the underlying training data and the limited diversity of pseudo-labels~\cite{yuan2021closer,gao2021learning}.

In contrast, we focus on the zero-shot setting, which aims to achieve strong generalization by harnessing the multimodal understanding and reasoning abilities of pre-trained Large Vision-Language Models (LVLMs)~\cite{radford2021clip,li2023blip2,Xiao_2026_CVPR,chen2024videollm,tang2026charts}, without any task-specific fine-tuning~\cite{zhang2025visual,zhang2026leveraging,zhang2024diff,xu2025zeroshotvmr,xu2024vtggpt}.

\subsection{Zero-shot Video Temporal Grounding}

Zero-shot VTG methods~\cite{luo2024zeroshot,xu2024vtggpt,zheng2024tfvtg,xu2025zeroshotvmr,lee2025tag} leverage pre-trained VLMs~\cite{radford2021clip,li2023blip2} and LVLMs~\cite{zhang2023videollama,wang2024internvideo} as feature extractors to obtain embeddings for both queries and video frames.
In these frameworks, the target event is localized by computing similarity scores between textual and visual embeddings and merging temporally adjacent high-scoring frames into temporal proposals~\cite{luo2024zeroshot,zheng2024tfvtg}.
Luo~\etal~\cite{luo2024zeroshot}, who first introduced the zero-shot VTG task, identify target events via similarity matching in a shared feature space, directly inferring temporal boundaries from matched frames.
Recognizing that queries often describe composite events involving multiple sub-events, Zheng~\etal~\cite{zheng2024tfvtg} decompose queries into sub-event sequences, match each to corresponding visual segments, and aggregate the results to derive the final temporal span.
More recently, Jeon~\etal~\cite{jeon2026granalign} address the granularity mismatch between queries and video content by rewriting queries at multiple granularity levels and aligning each rewritten query with corresponding video captions via CLIP-based cosine similarity. Mechanism-level explanations clarify how networks transform and route information for interpretable multimodal reasoning~\cite{Cai_2023_ICCV,Cai_2024_CVPR,Cai_2026_CVPR,caitoward}.
This line of work provides a valuable foundation, which we complement by using the spectrum of query-conditioned frame relations to decide when temporal reasoning is needed.

Despite these advances, all above methods apply a fixed processing pipeline to every query regardless of its complexity.
In contrast, our method introduces difficulty-adaptive routing that automatically distinguishes simple from complex queries and activates structured temporal reasoning only when needed.

\section{Method}
\label{sec:method}

Given an untrimmed video and a natural language query, video temporal grounding (VTG) aims to predict the start and end timestamps of the segment most semantically relevant to the query.

\begin{figure*}[t]
\centering
\includegraphics[width=\textwidth]{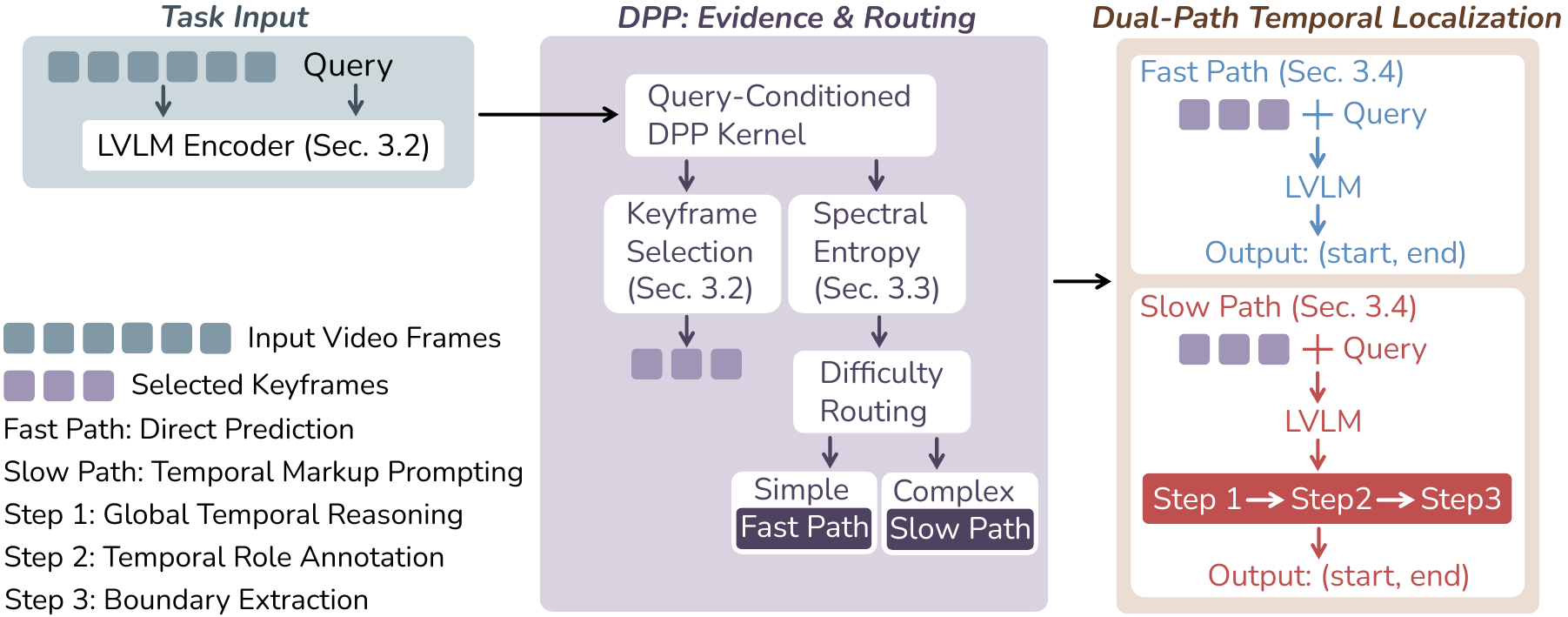}
\caption{Overview of the \method{} pipeline. The LVLM encoder denotes the vision encoder and text encoder used to extract frame features and query features, respectively. \method{} then (1) selects diverse, query-relevant keyframes via a DPP kernel, (2) routes each query to a fast or slow path based on spectral entropy, and (3) performs temporal localization through either direct prediction or structured reasoning.}
\label{fig:pipeline}
\end{figure*}

To bridge the reasoning gap of existing feature-matching methods, we propose \method{} (\cref{fig:pipeline}), which activates the reasoning capabilities of LVLMs through three tightly coupled stages.
We first construct a query-conditioned DPP kernel that jointly encodes frame-query relevance and inter-frame visual-temporal diversity, then greedily select keyframes with an adaptive stopping criterion that automatically determines the frame budget for each query (\cref{sec:kernel}).
The spectral entropy of this kernel, which reflects the relational structure among event stages, then serves as a difficulty indicator that routes each query to either direct localization or structured reasoning (\cref{sec:spectral}).
Based on this decision, the selected keyframes are passed to either direct localization or structured temporal reasoning via our Temporal Markup Prompting (\cref{sec:dual_path}).
We introduce necessary background on DPP in \cref{sec:preliminary}.

\subsection{Preliminaries: Determinantal Point Processes}
\label{sec:preliminary}

Selecting a subset that balances element importance and inter-element diversity is computationally intractable~\cite{kuo1993maxdiversity}: the diversity criterion couples every element to every other, so each element's value depends on the composition of the entire selected set, requiring in principle an exhaustive search over all possible subsets.
A Determinantal Point Process (DPP)~\cite{kulesza2012dpp} provides an efficient solution: it encodes both importance and diversity into a single kernel matrix, from which a near-optimal subset can be obtained via greedy selection with provable approximation guarantees.

Specifically, given a ground set of $M$ elements, a DPP captures element relationships through a kernel matrix $\bm{L} \in \mathbb{R}^{M \times M}$, whose diagonal entries $L_{ii}$ encode element importance and off-diagonal entries $L_{ij}$ encode pairwise similarity.
For any subset $S$, the determinant $\det(\bm{L}_S)$ of the corresponding submatrix scores its quality: it measures the ``volume'' spanned by the selected elements, which is large when each element is important (large diagonal) and the elements are diverse (small off-diagonal).
Exact maximization of $\det(\bm{L}_S)$ over all subsets is intractable, but a greedy algorithm that iteratively selects the element with the largest \emph{marginal gain} $\Delta(e \mid S)$ achieves a near-optimal solution with provable guarantees~\cite{kulesza2012dpp}.
The marginal gain measures how much adding element $e$ to the current set $S$ increases the log-determinant:
\begin{equation}
\label{eq:marginal_gain}
\Delta(e \mid S) = \log\det(\bm{L}_{S \cup \{e\}}) - \log\det(\bm{L}_S).
\end{equation}
The process repeats until the desired number of elements is reached.

\subsection{Preparing Temporal Evidence via Query-Conditioned DPP}
\label{sec:kernel}

Effective CoT reasoning requires selecting keyframes that are both relevant to the query and mutually diverse, covering the target event while minimizing redundancy.
Since jointly optimizing both criteria requires evaluating all possible frame subsets, which is exponential in the number of frames, we leverage the DPP (\cref{sec:preliminary}) and design a \emph{query-conditioned} variant that greedily selects diverse yet query-relevant keyframes.
Specifically, it consists of two steps: (1)~constructing a query-conditioned kernel that jointly encodes frame-query relevance and inter-frame visual-temporal similarity, and (2)~greedily selecting keyframes from this kernel with an adaptive stopping criterion.

\noindent\textbf{Query-Conditioned Kernel Construction.}
To jointly capture frame-query relevance and inter-frame similarity within a single optimization objective, we construct a query-conditioned kernel matrix $\tilde{\bm{L}}$.
We first build a base similarity kernel $\bm{L}$ that models visual-temporal relationships among frames, then weight it by query-dependent relevance scores to obtain $\tilde{\bm{L}}$.

Specifically, given frame features $\bm{F} = [\bm{f}_1, \dots, \bm{f}_M]$ from the vision encoder of a pretrained LVLM and a text feature $\bm{q}$ from the same LVLM's text encoder, the base similarity kernel $L_{ij} = k_v(\bm{f}_i, \bm{f}_j) \cdot k_t(t_i, t_j)$ combines a visual Gaussian kernel $k_v$ and a temporal Gaussian kernel $k_t$.
The product form preserves visually similar frames that are far apart in time, allowing different stages of the same event to be retained.
To account for query relevance, we further weight each frame by its relevance score $w_i = \sigma(\bm{f}_i^\top \bm{q} / \tau)$, where $\sigma$ is the sigmoid function and $\tau$ is a temperature parameter, and construct the query-conditioned kernel $\tilde{L}_{ij} = \sqrt{w_i} \; L_{ij} \; \sqrt{w_j}$.
The sigmoid maps frame-query similarity into $[0, 1]$, ensuring that query-irrelevant frames receive near-zero weight and are effectively suppressed in the kernel.
The resulting $\tilde{\bm{L}}$ jointly encodes frame-query relevance (via $w_i$) and inter-frame diversity (via $L_{ij}$), enabling our query-conditioned DPP to select frames that satisfy both criteria simultaneously.

\noindent\textbf{Greedy Keyframe Selection with Adaptive Budget.}
After obtaining $\tilde{\bm{L}}$, we greedily select keyframes by iteratively choosing the frame with the largest marginal gain $\Delta(e \mid S)$ (\cref{eq:marginal_gain}).
Since $\tilde{L}_{ij} = \sqrt{w_i}\,L_{ij}\,\sqrt{w_j}$, by the multiplicative property of determinants~\cite{kulesza2012dpp} we have $\det(\tilde{\bm{L}}_S) = (\prod_{i \in S} w_i)\,\det(\bm{L}_S)$.
Substituting into \cref{eq:marginal_gain} yields:
\begin{align}
  \Delta(e \mid S)
  &= \log\det(\tilde{\bm{L}}_{S \cup \{e\}}) - \log\det(\tilde{\bm{L}}_S) \notag \\
  &= \log\!\bigl[\textstyle\prod_{i \in S \cup \{e\}} w_i \cdot \det(\bm{L}_{S \cup \{e\}})\bigr] - \log\!\bigl[\textstyle\prod_{i \in S} w_i \cdot \det(\bm{L}_S)\bigr] \notag \\
  &= \underbrace{\log\det(\bm{L}_{S \cup \{e\}}) - \log\det(\bm{L}_S)}_{\text{diversity}} + \underbrace{\log w_e}_{\text{query relevance}},
  \label{eq:conditioned_gain}
\end{align}
Intuitively, \cref{eq:conditioned_gain} shows that each selected frame balances two complementary objectives: the diversity term favors frames dissimilar to those already selected, while the relevance term favors frames closely related to the query.

As different events may span different numbers of stages and durations, the number of keyframes needs to vary across queries.
In our method, we let the greedy process itself decide when to stop.
The key observation is that $\Delta_k$ typically decreases as the greedy process proceeds: early frames capture the most informative content with large gains, while later frames contribute progressively less.
We therefore stop when newly added frames contribute negligible information, \ie, when the relative gain drops below a threshold: $\Delta_k / \Delta_1 < \varepsilon$, where $\Delta_1$ is the first-step gain and $\varepsilon$ is a hyperparameter.
This way, the number of selected keyframes automatically adapts to each (video, query) pair.
The greedy process outputs $K$ keyframes paired with their timestamps as input for the LVLM.

\subsection{Spectral Analysis for Difficulty Routing}
\label{sec:spectral}

After obtaining the adaptive keyframe budget for each event (\cref{sec:kernel}), we next route each query to either direct localization or structured reasoning.
A straightforward approach is to use the frame count itself as a proxy for difficulty, since events requiring more keyframes seem to be harder.
However, this heuristic is unreliable: ``person walks from the living room to the kitchen'' may yield many keyframes because the scene changes significantly across rooms, yet the action is simple and needs no reasoning.
Conversely, ``person picks up a phone and answers it'' may need only a few keyframes, yet involves a causal chain of distinct sub-actions that demands structured reasoning.
What truly distinguishes complex queries is not the number of frames but the \emph{relational structure} among event stages, including temporal ordering, causal transitions, and conditional dependencies (\cref{sec:intro}).

Since our query-conditioned kernel $\tilde{\bm{L}}$ already encodes such relational structure among all frames, a natural question arises: can we read off the reasoning complexity directly from $\tilde{\bm{L}}$?
Indeed, the eigenspectrum of a positive semidefinite kernel reveals how many independent components the data contains~\cite{kulesza2012dpp}: a few dominant eigenvalues indicate that the data has few independent modes, while many significant eigenvalues signal a richer multi-modal structure.
Because $\tilde{\bm{L}}$ is conditioned on the query, frames unrelated to the query receive near-zero weight and are effectively suppressed in the kernel, so the eigenspectrum reflects the complexity of the query-relevant event structure rather than the general visual diversity of the video.
Each dominant eigenvalue therefore corresponds to a group of query-relevant yet mutually dissimilar frames---intuitively, a distinct event stage; a simple event yields one or two dominant eigenvalues, while a complex multi-stage event spreads energy across many (\cref{fig:spectral_contrast}).
Building on this, we quantify the number of dominant eigenvalues via the spectral entropy of $\tilde{\bm{L}}$ to determine whether structured reasoning is needed, from a single spectral decomposition at negligible cost.

\definecolor{clrSimple}{HTML}{D06C9D}
\definecolor{clrComplex}{HTML}{FDE8D3}

\pgfplotsset{
  spectralbar/.style={
    ybar,
    bar width=7pt,
    width=\linewidth,
    height=4.5cm,
    ymin=0, ymax=10,
    xtick={1,2,3,4,5,6,7},
    xticklabels={1,2,3,4,5,6,7},
    ytick={0,2,4,6,8},
    ticklabel style={font=\sffamily\scriptsize},
    label style={font=\sffamily\small},
    title style={font=\sffamily\small, at={(0.5,0.98)}, anchor=south},
    enlarge x limits=0.15,
    axis x line*=bottom,
    axis y line*=left,
    ytick align=outside,
    xtick style={draw=none},
    major tick length=2pt,
    ylabel={Eigenvalue $\lambda_i$},
    xlabel={Index $i$},
    nodes near coords={},%
  }
}

\begin{figure}[t]
  \centering\vspace{-2mm}
  \begin{minipage}[t]{0.48\linewidth}
    \centering
    \begin{tikzpicture}
      \begin{axis}[
        spectralbar,
        title={\emph{``A person opens a door''}},
      ]
      \addplot[fill=clrSimple, draw=clrSimple!80!black] coordinates {
        (1,8.5) [] (2,0.4) [] (3,0.2) [] (4,0.1) [] (5,0.05) [] (6,0.03) [] (7,0.02) []
      };
      \node[font=\sffamily\scriptsize, fill=clrSimple!15, rounded corners=2pt, inner sep=3pt, anchor=north]
        at ([yshift=-3pt]current axis.south) {%
        $H_{\mathrm{sp}}\!\approx\!0.21$ \;\; $\to$ \textbf{Fast path}};
      \end{axis}
    \end{tikzpicture}
    \label{fig:spectral_simple}
  \end{minipage}
  \hfill
  \begin{minipage}[t]{0.48\linewidth}
    \centering
    \begin{tikzpicture}
      \begin{axis}[
        spectralbar,
        title={\emph{``A gymnast runs, flips, and lands''}},
      ]
      \addplot[fill=clrComplex, draw=clrComplex!40!black, line width=0.4pt] coordinates {
        (1,5.2) [] (2,4.1) [] (3,3.3) [] (4,0.3) [] (5,0.15) [] (6,0.08) [] (7,0.04) []
      };
      \node[font=\sffamily\scriptsize, fill=clrComplex!50, rounded corners=2pt, inner sep=3pt, anchor=north]
        at ([yshift=-3pt]current axis.south) {%
        $H_{\mathrm{sp}}\!\approx\!0.65$ \;\; $\to$ \textbf{Slow path}};
      \end{axis}
    \end{tikzpicture}
    \label{fig:spectral_complex}
  \end{minipage}\vspace{-2mm}
  \caption{%
    \textbf{Spectral contrast between simple and complex queries.}
    Eigenvalue spectra of the query-conditioned kernel $\tilde{\bm{L}}$ for a Charades-STA query (left) and an ActivityNet Captions query (right).
    A simple query (left) concentrates energy in $\lambda_1$ (one event stage), yielding low $H_{\mathrm{spectral}}$ $\to$ \textbf{Fast path}.
    A complex query (right) spreads energy across $\lambda_1$--$\lambda_3$ (multiple event stages), yielding high $H_{\mathrm{spectral}}$ $\to$ \textbf{Slow path}.
  }
  \label{fig:spectral_contrast}
\end{figure}
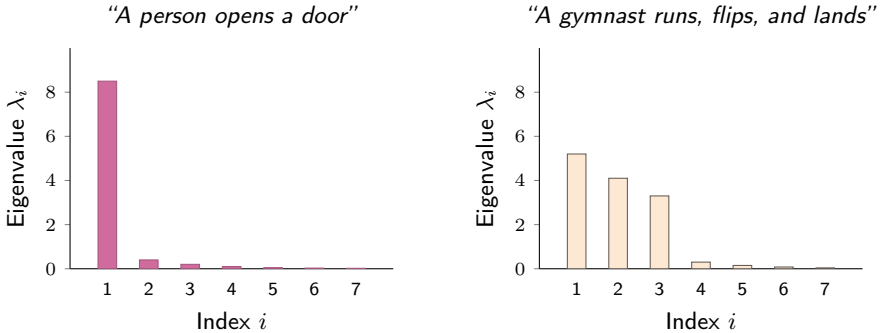

\noindent\textbf{Spectral Entropy as Difficulty Indicator.}
As discussed above, the number of dominant eigenvalues serves as a reliable indicator that distinguishes simple from complex events.
However, directly counting dominant eigenvalues requires choosing an energy threshold, and different thresholds can lead to inconsistent counts for borderline cases, making the indicator unreliable.
To avoid this, we normalize the eigenvalue spectrum into a probability distribution and compute its entropy, whose information-theoretic meaning is the \emph{effective number} of dominant eigenvalues~\cite{roy2007effective}, providing a smooth, information-theoretic difficulty measure that captures the complexity of the query-relevant temporal structure.

Specifically, we perform eigendecomposition of $\tilde{\bm{L}}$ to obtain its eigenvalues $\{\lambda_i\}_{i=1}^{M}$, normalize them into $q_i = \lambda_i / \sum_j \lambda_j$, and compute the spectral entropy $H_{\mathrm{spectral}} = -\tfrac{1}{\log M}\sum_{i=1}^{M} q_i \log q_i \;\in\; [0,\,1]$.
$H_{\mathrm{spectral}}$ close to 0 indicates that only one or two eigenvalues dominate (few event stages, simple query); $H_{\mathrm{spectral}}$ close to 1 indicates that many eigenvalues carry comparable energy (many event stages, complex query).

\noindent\textbf{Difficulty-Aware Routing.}
Based on $H_{\mathrm{spectral}}$, we route each query to one of two reasoning paths: if $H_{\mathrm{spectral}} \leq \theta$, the query is assigned to the \textbf{Fast} path, which directly prompts the LVLM to predict start and end timestamps from the selected keyframes, sufficing for simple events; otherwise ($H_{\mathrm{spectral}} > \theta$), it is routed to the \textbf{Slow} path, which activates our Temporal Markup Prompting (\cref{sec:dual_path}), a structured multi-step reasoning process designed for complex multi-stage events. Here $\theta$ is a hyperparameter.

\subsection{Dual-Path Temporal Localization}
\label{sec:dual_path}

At this point, we have determined both the keyframe set and the reasoning path for each query.
For simple queries ($H_{\mathrm{spectral}} \leq \theta$), the event structure is already clear from the selected keyframes, so the LVLM only needs to identify the matching segment.
For complex queries ($H_{\mathrm{spectral}} > \theta$), however, the event spans multiple stages with temporal and causal dependencies that cannot be resolved by visual matching alone, requiring the LVLM to reason about how the stages relate before localizing the boundaries.
We design two corresponding paths to match these distinct requirements.

\noindent\textbf{Fast Path: Direct Prediction.}
For simple queries routed to the Fast path, we provide the $K$ selected keyframes with their timestamps to the LVLM along with a task-specific instruction that describes the temporal grounding objective and specifies the output format.
The LVLM directly outputs the start and end timestamps $(t_s, t_e)$ without any intermediate reasoning steps.

\noindent\textbf{Slow Path: Temporal Markup Prompting.}
For complex queries, directly prompting the LVLM to output timestamps can be suboptimal, as the model needs to implicitly reason about multiple actions, their temporal ordering, and their correspondence to specific frames in a single step.
Moreover, generic Chain-of-Thought prompting (\eg ``let's think step by step'') provides no task-specific guidance and risks producing verbose, unfocused reasoning.
Inspired by recent work showing that structured, task-specific prompting significantly outperforms open-ended reasoning for spatial and temporal tasks~\cite{chen2024spatialvlm,huang2024vtimellm}, we design \emph{Temporal Markup Prompting}~(TMP), a structured reasoning procedure specifically tailored for temporal grounding.
Unlike generic CoT, TMP prescribes a fixed three-step procedure with well-defined intermediate outputs at each step, ensuring that every step is directed toward temporal localization.
The three steps are executed sequentially within a single generation:

\textit{Step~1: Global Temporal Reasoning.}
Before looking at any individual frame, the model is prompted to break the query into its key actions, identify how they relate in time (sequential, causal, conditional, \etc), and outline where the event roughly falls on the timeline.
This step ensures that the model forms a global picture of the event first, so that subsequent per-frame judgments are anchored in temporal context rather than made in isolation.

\textit{Step~2: Per-Frame Temporal Role Annotation.}
Guided by the global analysis from Step~1, we prompt the model to assign every keyframe a temporal role from the ordered set $\{\textsc{before}, \textsc{start}, \textsc{during}, \textsc{end}, \textsc{after}\}$.
Two constraints enforce structured output: (i)~every keyframe must receive a label, preventing the model from skipping frames or relying on implicit assumptions; (ii)~labels must be monotonically non-decreasing along the timeline (\textsc{before} $\to$ \textsc{start} $\to$ \textsc{during} $\to$ \textsc{end} $\to$ \textsc{after}), enforcing temporal consistency.
Note that the diversity term in \cref{sec:kernel} ensures keyframes extend beyond event boundaries, providing context for reliable annotation.

\textit{Step~3: Boundary Extraction.}
The start timestamp $t_s$ is taken from the first frame labeled \textsc{start}, and $t_e$ from the last frame labeled \textsc{end}.

\section{Experiments}
\label{sec:experiments}

\subsection{Experimental Setups}
\label{sec:setup}

\noindent\textbf{Datasets.}
For fair comparison, we follow~\cite{zheng2024tfvtg} to conduct experiments on two benchmark datasets: ActivityNet Captions~\cite{krishna2017activitynet} and Charades-STA~\cite{gao2017tall}.
ActivityNet Captions contains 20K videos originally collected for video captioning, with 37,417/17,505/17,031 video-query pairs in the train/val\_1/val\_2 splits.
Following previous works~\cite{luo2024zeroshot,zheng2024tfvtg}, we report results on the val\_2 split.
Charades-STA is constructed based on the Charades dataset~\cite{sigurdsson2016hollywood} and includes 12,408/3,720 video-query pairs in the train/test splits.
We report performance on the test split.

\noindent\textbf{Evaluation Metrics.}
We follow the evaluation metrics ``R@m'' and ``mIoU'' used in previous works~\cite{luo2024zeroshot,zheng2024tfvtg}.
The metric ``R@x'' represents the percentage of predictions whose Intersection over Union (IoU) exceeds ``x'', while ``mIoU'' measures the average IoU across all predictions.

\noindent\textbf{Implementation Details.}
We adopt LLaVA-1.6-7B~\cite{liu2024llavanext} as our vision-language backbone.
Each input video is downsampled to 3 FPS following TFVTG~\cite{zheng2024tfvtg} to ensure consistent evaluation.
For the query-conditioned DPP kernel (\cref{sec:kernel}), we set Gaussian kernel bandwidths $\sigma_v{=}0.5$ for visual similarity and $\sigma_t{=}2.0$ for temporal distance, with temperature $\tau{=}0.1$ in the relevance weight; all three are selected via grid search on the Charades-STA validation split.
The adaptive stopping threshold is set to $\varepsilon{=}0.05$, and the routing threshold to $\theta{=}0.45$.
During inference, the LVLM uses greedy decoding with a maximum of 512 tokens.
All experiments are conducted on a single NVIDIA A100 80\,GB GPU.
\subsection{Main Results}
\label{sec:main_results}

We follow TFVTG~\cite{zheng2024tfvtg} and conduct experiments on the Charades-STA~\cite{gao2017tall} and ActivityNet Captions~\cite{krishna2017activitynet} datasets under both the identically distributed (IID) setting and the out-of-distribution (OOD) setting.

\begin{table*}[t]
    \centering
    \caption{Results under the IID setting on Charades-STA~\cite{gao2017tall} and ActivityNet Captions~\cite{krishna2017activitynet}. Best zero-shot results are in \textbf{bold}, second best are \underline{underlined}.}

    \scriptsize
    \begin{tabular}{l|c|cccc|cccc}
    \toprule
     \multirow{2}{*}{Method} & \multirow{2}{*}{Setting} & \multicolumn{4}{c|}{Charades-STA~\cite{gao2017tall}}  &\multicolumn{4}{c}{ActivityNet Captions~\cite{krishna2017activitynet}} \\
           &  & R@0.3 & R@0.5 & R@0.7 & mIoU &R@0.3 & R@0.5 & R@0.7 & mIoU \\
    \midrule
    GroundingGPT~\cite{li2024groundgpt}&\multirow{6}{*}{Fully}& - & 29.6 &11.9 & -&-&-&-&-\\
    VTimeLLM-13B~\cite{huang2024vtimellm} & &55.3 &34.3 &14.7 &34.6& 44.8 &29.5 &14.2 &31.4 \\
       2D-TAN~\cite{zhang2020learning}& & - & 39.81 & 23.25 & -& 58.75 & 44.05 & 27.38 & - \\
        EMB~\cite{huang2022video}&  & 72.50 & 58.33 & 39.25 & 53.09& 64.13 & 44.81 & 26.07 & 45.59\\
        MGSL-Net~\cite{liu2022memory}& & - & 63.98 & 41.03 & -& - & 51.87 & 31.42 & -\\
        EaTR~\cite{jang2023EaTR}& & - & 68.47 & 44.92 & -& - & 58.18 & 37.64 & -\\
    \midrule
        CRM~\cite{huang2021cross}&\multirow{4}{*}{Weakly}& 53.66 & 34.76 & 16.37 &- & 55.26 & 32.19 &-&-  \\
        CNM~\cite{CNM_2022_AAAI}&&60.39 & 35.43 & 15.45& -&55.68 & 33.33 &-& -\\
        CPL~\cite{CPL_2022_CVPR}& & 66.40 &49.24& 22.39 & -& 55.73 &31.37&- & -\\
         Huang~\etal~\cite{huang2023weakly} &  & 69.16& 52.18& 23.94 &45.20 & 58.07 &36.91& - & 41.02 \\
    \midrule
        Gao~\etal~\cite{gao2021learning}&\multirow{5}{*}{Unsup.}& 46.69 & 20.14 & 8.27 & -& 46.15 & 26.38 & 11.64 & - \\
        PSVL~\cite{nam2021zero}& & 46.47 & 31.29& 14.17 &31.24 & 44.74 & 30.08 & 14.74 & 29.62\\
        PZVMR~\cite{wang2022prompt}& &   46.83 & 33.21 & 18.51 & 32.62& 45.73 & 31.26 & 17.84 & 30.35 \\
        Kim~\etal~\cite{kim2023language}& & 52.95 & 37.24 & 19.33 & 36.05& 47.61 & 32.59 & 15.42 & 31.85 \\
        SPL~\cite{zheng2023SPL} & & 60.73 & 40.70 & 19.62 & 40.47 & 50.24 & 27.24 & 15.03 & 35.44\\
    \midrule
    VideoChat-7B~\cite{li2023videochat} &\multirow{10}{*}{Zero-shot} &9.0& 3.3& 1.3 &6.5& 8.8& 3.7& 1.5 &7.2 \\
    VideoLLaMA-7B~\cite{zhang2023videollama}& &10.4 &3.8 &0.9 &7.1 &6.9 &2.1 &0.8 &6.5 \\
    VideoChatGPT-7B~\cite{maaz2023videochatgpt}& &20.0 &7.7 &1.7 &13.7& 26.4& 13.6& 6.1& 18.9\\
    Luo~\etal~\cite{luo2024zeroshot}& & 56.77 & 42.93& 20.13& 37.92 &48.28& 27.90& 11.57& 32.37\\
    GranAlign~\cite{jeon2026granalign}&&59.1 &39.6 &22.7 &38.0 &50.3 &\textbf{34.0} &\underline{16.5} &33.1\\
    VTG-GPT~\cite{xu2024vtggpt} & & 59.48& 43.68& 25.94& 39.81 &47.13& 28.25& 12.84 &30.49\\
    VTIMECOT~\cite{zhang2025vtimecot}&&66.96 &38.79 &20.83 &43.41 &- &- &- &-\\
    TFVTG~\cite{zheng2024tfvtg} &  &67.04 &\underline{49.97} & 24.32 & 44.51 & 49.34 &	27.02	& 13.39 &34.10 \\
    TAG~\cite{lee2025tag}&&\underline{67.82} &48.58 &\underline{26.67} &\underline{45.69} &\underline{51.88} &28.91 &15.07 &\underline{36.55}\\
    \rowcolor[HTML]{f3f7fc}
    Ours&&\textbf{70.98} &\textbf{52.04} &\textbf{29.45} &\textbf{48.93} &\textbf{54.90} &\underline{32.14} &\textbf{18.11} &\textbf{39.89}\\
    \bottomrule
    \end{tabular}
    \label{tab:iid}

\end{table*}

\noindent\textbf{IID Setting.}
As shown in \cref{tab:iid}, DART achieves state-of-the-art zero-shot performance on both datasets, surpassing the previous best zero-shot method TAG~\cite{lee2025tag} by $+3.24$ mIoU on Charades-STA and $+3.34$ mIoU on ActivityNet Captions.
Notably, DART also outperforms several weakly supervised methods (\eg, CPL~\cite{CPL_2022_CVPR}) and narrows the gap with fully supervised approaches.
We attribute this to the combination of query-conditioned DPP keyframe selection, which provides more informative visual input, and difficulty-adaptive routing, which applies structured CoT reasoning only when needed.

\begin{table*}[t]
    \centering
    \scriptsize
        \caption{Results under the OOD setting on the Charades-STA~\cite{gao2017tall} and ActivityNet Captions~\cite{krishna2017activitynet} datasets. }
    \resizebox{\textwidth}{!}{
    \begin{tabular}{lc|ccc|ccc|ccc|ccc}
    \toprule
         & \multirow{3}{*}{Method} & \multicolumn{6}{c|}{Charades-STA~\cite{gao2017tall}} & \multicolumn{6}{c}{ActivityNet Captions~\cite{krishna2017activitynet}} \\
         & & \multicolumn{3}{c|}{OOD-1} & \multicolumn{3}{c|}{OOD-2} & \multicolumn{3}{c|}{OOD-1} & \multicolumn{3}{c}{OOD-2} \\
         & & R@0.5 & R@0.7 & mIoU & R@0.5 & R@0.7 & mIoU & R@0.5 & R@0.7 & mIoU & R@0.5 & R@0.7 & mIoU \\
    \midrule
    \multirow{5}{*}{\rotatebox[origin=c]{90}{Fully}}
        &LGI~\cite{LGI} & 42.1 & 18.6 & 41.2 & 35.8 & 13.5 & 37.1 & 16.3 & 6.2 & 22.2 & 11.0 & 3.9 & 17.3 \\
        &2D-TAN~\cite{zhang2020learning} & 27.1 & 13.1 & 25.7 & 21.1 & 8.8 & 22.5 & 16.4 & 6.6 & 23.2 & 11.5 & 3.9 & 19.4 \\
        &MMN~\cite{wang2021unsupervised} & 31.6 & 13.4 & 33.4 & 27.0 & 9.3 & 30.3 & 20.3 & 7.1 & 26.2 & 14.1 & 5.2 & 20.6 \\
        &VDI~\cite{luo2023vid} & 25.9 & 11.9 & 26.7 & 20.8 & 8.7 & 22.0 & 20.9 & 7.1 & 27.6 & 14.3 & 5.2 & 23.7 \\
        &DCM~\cite{yang2021deconfounded} & 44.4 & 19.7 & 42.3 & 38.5 & 15.4 & 39.0 & 18.2 & 7.9 & 24.4 & 12.9 & 4.8 & 20.7 \\
    \midrule
    \multirow{2}{*}{\rotatebox[origin=c]{90}{W.}}
        &CNM~\cite{CNM_2022_AAAI} & 9.9 & 1.7 & 21.6 & 6.1 & 0.5 & 16.6 & 6.1 & 0.4 & 21.0 & 2.5 & 0.1 & 16.8 \\
        &CPL~\cite{CPL_2022_CVPR} & 29.9 & 8.5 & 32.2 & 24.9 & 6.3 & 30.5 & 4.7 & 0.4 & 21.1 & 2.1 & 0.2 & 17.7 \\
    \midrule
    \multirow{2}{*}{\rotatebox[origin=c]{90}{U.}}
        &PSVL~\cite{nam2021zero} & 3.0 & 0.7 & 8.2 & 2.2 & 0.4 & 6.8 & - & - & - & - & - & - \\
        &PZVMR~\cite{wang2022prompt} & - & 8.6 & 25.1 & - & 6.5 & 28.5 & - & 4.4 & 28.3 & - & 2.6 & 19.1 \\
    \midrule
    \multirow{4}{*}{\rotatebox[origin=c]{90}{ZS}}
        &Luo~\etal~\cite{luo2024zeroshot} & 40.3 & 18.2 & 38.2 & 38.9 & 17.0 & 37.8 & 18.4 & 6.8 & 21.1 & 18.6 & 7.4 & 20.6 \\
        &TFVTG~\cite{zheng2024tfvtg} &\underline{45.9} &20.8&43.0&43.8&20.0&42.6&20.4&11.2&31.7&18.5&10.0&30.3 \\
        &TAG~\cite{lee2025tag} &45.3 &\underline{23.2} &\underline{44.7} &\underline{44.1} &\underline{22.0} &\underline{44.6} &\underline{28.5} &\underline{14.7} &\underline{36.2} &\underline{28.3} &\underline{14.5} &\underline{36.1}\\
        \rowcolor[HTML]{f3f7fc} &Ours &\textbf{48.7} &\textbf{25.8} &\textbf{48.1} &\textbf{47.2} &\textbf{24.5} &\textbf{47.8} &\textbf{31.8} &\textbf{17.9} &\textbf{39.7} &\textbf{31.6} &\textbf{17.6} &\textbf{39.5}\\
    \bottomrule
    \end{tabular}
    }
    \label{tab:2}
\end{table*}

\begin{table*}[t]
    \centering
    \scriptsize
    \caption{Results on Charades-CD~\cite{yuan2021closer} (OOD split) and Charades-CG~\cite{li2022compositional} (novel text).}

    \begin{tabular}{lc|ccc|ccc|ccc}
    \toprule
         &\multirow{3}{*}{Method}  & \multicolumn{3}{c|}{Charades-CD~\cite{yuan2021closer}} &\multicolumn{6}{c}{Charades-CG~\cite{li2022compositional}}  \\
         & & \multicolumn{3}{c|}{Test-OOD} & \multicolumn{3}{c|}{Novel-composition} & \multicolumn{3}{c}{Novel-word} \\
         & & R@0.3 & R@0.5 & R@0.7 & R@0.5 & R@0.7 & mIoU & R@0.5 & R@0.7 & mIoU \\
    \midrule
    \multirow{5}{*}{\rotatebox[origin=c]{90}{Fully}}
        &2D-TAN~\cite{zhang2020learning}  & 43.45 & 30.77 & 11.75 & 30.91 & 12.23 & 29.75 & 29.36 & 13.21 & 28.47 \\
        &TSP-PRL~\cite{wu2020tree}  & 31.93 & 19.37 & 6.20 & 16.30 & 2.04 & 13.52 & 14.83 & 2.61 & 14.03 \\
        &SCDM~\cite{yuan2019semantic}   & 52.38 & 41.60 & 22.22 & 27.73 & 12.25 & 30.84 &- & -&- \\
        &VISA~\cite{li2022compositional}   & - & - & - & 45.41 & 22.71 & 42.03 & 42.35 & 20.88 & 40.18 \\
        &DeCo~\cite{yang2023deco}  & - & - & - & 47.39 & 21.06 & 40.70 & - & -  & -  \\
    \midrule
    \multirow{2}{*}{\rotatebox[origin=c]{90}{W.}}
        &WSSL~\cite{duan2018weakly}  & 35.86 & 23.67 & 8.27 & 3.61 & 1.21 & 8.26 & 2.79 & 0.73 & 7.92 \\
        &CPL~\cite{CPL_2022_CVPR}   & - & - & - & 39.11 & 15.60 & 35.53 &  45.90 &22.88 &-   \\
    \midrule
    \multirow{1}{*}{\rotatebox[origin=c]{90}{U.}}
        &SPL~\cite{zheng2023SPL}  & 62.96 & 38.25 & 15.53 & - & -& -&- &-&-\\
    \midrule
    \multirow{4}{*}{\rotatebox[origin=c]{90}{ZS}}
        &Luo~\etal~\cite{luo2024zeroshot}  &- & -& -& 40.27 &16.27 & -&45.04& 21.44 & -\\
        &TFVTG~\cite{zheng2024tfvtg}  &65.07&49.24 &23.05&\underline{43.84} &18.68 & 40.19&\underline{56.26}&28.49&46.90\\
        &TAG~\cite{lee2025tag} &\underline{67.94} &\underline{50.19} &\underline{27.14} &43.55 &\underline{21.30} &\underline{41.95} &52.37 &\underline{32.66} &\underline{47.86}\\
         \rowcolor[HTML]{f3f7fc} &Ours &\textbf{70.41} &\textbf{53.75}  &\textbf{30.45}  &\textbf{46.45}  &\textbf{24.11}  &\textbf{45.30}  &\textbf{55.61}  &\textbf{35.43}  &\textbf{51.09} \\
    \bottomrule
    \end{tabular}
    \label{tab:ood_cd}

\end{table*}

\begin{table*}[t]
    \centering
    \begin{minipage}[t]{0.48\textwidth}
        \centering
        \caption{Cross-dataset: train on ActivityNet Captions, test on Charades-STA.}
        \scriptsize
        \begin{tabular}{l|cccc}
        \toprule
             \multirow{2}{*}{Method} & R@1 & R@1 & R@5 & R@5 \\
             & R@0.5 & R@0.7 & R@0.5 & R@0.7 \\
        \midrule
             SCDM~\cite{yuan2019semantic} & 15.91 & 6.19 & 54.04 & 30.39 \\
             2D-TAN~\cite{zhang2020learning} & 15.81 & 6.30 & 59.06 & 31.53 \\
             Debias-TLL~\cite{bao2022learning} & 21.45 & 10.38 & 62.34 & 32.90 \\
             TFVTG~\cite{zheng2024tfvtg} & \underline{49.97} & \underline{24.32} &  \underline{83.50} & \underline{42.20}  \\
             \rowcolor[HTML]{f3f7fc}  Ours & \textbf{53.17} & \textbf{27.84} &  \textbf{87.43} & \textbf{46.97}  \\
        \bottomrule
        \end{tabular}
        \label{tab:cross}
    \end{minipage}%
    \hspace{1mm}
    \begin{minipage}[t]{0.48\textwidth}
        \centering
        \caption{Keyframe selection comparison on Charades-STA (IID).}
        \scriptsize
        \begin{tabular}{l|cccc}
        \toprule
             Method & R@0.3 & R@0.5 & R@0.7 & mIoU \\
        \midrule
             $K$-medoids          & 67.24 & 47.53 & 26.18 & 45.73 \\
             Uniform + re-rank    & 68.15 & 48.62 & 27.08 & 46.52 \\
             Top-$k$ + NMS        & 69.07 & 49.81 & 27.92 & 47.38 \\
        \midrule
             \rowcolor[HTML]{f3f7fc} DPP (Ours) & \textbf{70.98} & \textbf{52.04} & \textbf{29.45} & \textbf{48.93} \\
        \bottomrule
        \end{tabular}
        \label{tab:ablation_dpp}
    \end{minipage}
\end{table*}

\noindent\textbf{OOD Setting.}
Following TFVTG~\cite{zheng2024tfvtg}, we evaluate under moment-shifted splits where a random video of $p$ seconds is prepended to shift ground-truth locations (OOD-1/OOD-2, \cref{tab:2}), the distribution-shifted Charades-CD~\cite{yuan2021closer}, and the novel-text Charades-CG~\cite{li2022compositional} (\cref{tab:ood_cd}).
We also report the cross-dataset setting (\cref{tab:cross}), where supervised baselines are trained on ActivityNet Captions and tested on Charades-STA, while zero-shot methods are directly evaluated.
DART achieves state-of-the-art zero-shot performance across nearly all OOD metrics, demonstrating that difficulty-adaptive reasoning generalizes better than prior feature-matching methods~\cite{luo2024zeroshot} that primarily treat LVLMs as feature extractors.

\subsection{Ablation Studies}
\label{sec:ablation}

\begin{table*}[t]
    \centering
    \begin{minipage}[t]{0.56\textwidth}
        \centering
        \caption{Component ablation on Charades-STA (IID). Each row removes one component from the full \method{}.}
        \scriptsize
        \setlength{\tabcolsep}{3pt}
        \begin{tabular*}{\linewidth}{@{\extracolsep{\fill}}l|cccc@{}}
        \toprule
             Variant & R@0.3 & R@0.5 & R@0.7 & mIoU \\
        \midrule
              Ours & \textbf{70.98} & \textbf{52.04} & \textbf{29.45} & \textbf{48.93} \\
        \midrule
             w/o DPP        & 66.51 & 46.79 & 25.55 & 44.70 \\
             w/o Adaptive $K$ & 68.62 & 49.01 & 26.73 & 47.26 \\
             w/o TMP        & 68.14 & 48.65 & 25.92 & 46.13 \\
             Fast only      & 66.35 & 46.71 & 25.34 & 44.34 \\
             Slow only      & 68.71 & 49.80 & 26.78 & 47.53 \\
        \bottomrule
        \end{tabular*}
        \label{tab:ablation_component}
    \end{minipage}%
    \hfill
    \begin{minipage}[t]{0.40\textwidth}
        \centering
        \caption{Inference efficiency on Charades-STA. Time per query on a single A100 GPU.}
        \scriptsize
        \setlength{\tabcolsep}{3pt}
        \begin{tabular*}{\linewidth}{@{\extracolsep{\fill}}l|cc@{}}
        \toprule
             Method & Frames & Time (s) \\
        \midrule
             TFVTG~\cite{zheng2024tfvtg}    & 86 & 4.7 \\
             TAG~\cite{lee2025tag}           & 86 & 3.4 \\
        \midrule
             \method{} (Slow)     & 12 & 5.1 \\
              \method{} (Adaptive) & 12 & 3.9 \\
        \bottomrule
        \end{tabular*}
        \label{tab:efficiency}
    \end{minipage}
\end{table*}

\noindent\textbf{Component ablation.}
To verify the contribution of each component, we remove them individually and report results on Charades-STA (IID) in \cref{tab:ablation_component}.
First, replacing the query-conditioned DPP with uniform sampling (w/o DPP) causes the largest single-component drop ($-4.23$ mIoU), even though spectral routing and TMP remain active.
Notably, this variant only marginally outperforms Fast only ($44.70$ vs.\ $44.34$ mIoU), indicating that routing and structured reasoning provide little benefit when the underlying keyframes are uninformative, and that high-quality keyframe selection is a prerequisite for effective reasoning.
Second, replacing TMP with a generic chain-of-thought prompt (w/o TMP) reduces mIoU by $2.80$, confirming that task-specific structured reasoning outperforms unconstrained reasoning for temporal grounding.
Third, fixing the keyframe budget to $K{=}8$ instead of adaptive stopping (w/o Adaptive $K$) drops mIoU by $1.67$, showing that adapting the number of keyframes to each query is beneficial.
Finally, comparing the two routing extremes, Fast only ($44.34$), which skips reasoning entirely, vs.\ Slow only ($47.53$), which applies TMP to all queries without spectral routing, reveals that applying TMP uniformly is much better than skipping it entirely, yet the full adaptive routing ($48.93$) surpasses both, con\-firming that selectively applying reasoning to complex queries while protecting simple ones from over-reasoning yields the best trade-off.

\noindent\textbf{DPP vs.\ alternative keyframe selection.}
We compare our query-conditioned DPP against three alternatives in \cref{tab:ablation_dpp}. All variants share the same downstream pipeline (spectral routing + dual-path localization) and differ only in the keyframe selection strategy.
$K$-medoids clustering partitions frames into $K$ clusters by visual similarity and selects each medoid, maximizing spatial coverage but entirely ignoring query relevance (45.73 mIoU).
Uniform sampling with re-ranking first draws frames at equal intervals and then re-ranks them by query similarity to retain the top $K$; although query-aware, it is constrained by the quality of the initial uniformly-spaced candidate set (46.52 mIoU).
Top-$k$ similarity + NMS ranks all frames by query similarity, greedily selects the highest-scoring ones, and suppresses temporal neighbors within a fixed window; despite being the strongest baseline (47.38 mIoU), its greedy, independent selection still misses complementary event stages that a joint selection would capture.
DPP jointly optimizes relevance and diversity through the query-conditioned kernel (\cref{sec:kernel}), ensuring that selected keyframes are both highly relevant to the query and temporally diverse, yielding $+1.55$ mIoU over the best alternative.

\subsection{Analysis}
\label{sec:analysis}

\noindent\textbf{Spectral entropy reflects grounding difficulty.}
To validate that $H_{\mathrm{spectral}}$ is a reliable difficulty indicator, we randomly sample 1{,}000 queries from the ActivityNet Captions val set (200 per bin across five equal-width bins over $[0,\,1]$) and evaluate all queries using the \emph{Fast path only} (\ie, pure feature matching without structured reasoning), so that performance differences purely reflect query difficulty rather than the benefit of structured reasoning.
As shown in \cref{tab:perbin}, mIoU decreases monotonically from 49.2\% in the lowest bin to 27.4\% in the highest, confirming that higher spectral entropy corresponds to objectively harder grounding instances.
The sharp performance drop in high-entropy bins demonstrates that these queries cannot be resolved by feature matching alone and genuinely require the structured reasoning provided by the Slow path, validating $H_{\mathrm{spectral}}$ as an effective difficulty indicator for routing.

\begin{table}[t]
\centering
\caption{Per-bin mIoU \emph{vs.}\ $H_{\mathrm{spectral}}$ on 1{,}000 ActivityNet Captions val queries (200 per bin), evaluated using the Fast path only to ensure differences reflect query difficulty, not reasoning benefit. Higher entropy correlates with lower mIoU.}
\label{tab:perbin}
\setlength{\tabcolsep}{4pt}
\begin{tabular}{l|ccccc}
\toprule
$H_{\mathrm{spectral}}$ bin & [0,\,0.2) & [0.2,\,0.4) & [0.4,\,0.6) & [0.6,\,0.8) & [0.8,\,1.0] \\
\midrule
mIoU (\%) & 49.2 & 43.7 & 37.1 & 31.8 & 27.4 \\
\bottomrule
\end{tabular}
\end{table}

\noindent\textbf{Efficiency comparison.}
\cref{tab:efficiency} compares the inference cost of \method{} against zero-shot baselines on Charades-STA.
Despite processing far fewer frames per query (12 vs.\ 86), \method{} with adaptive routing achieves $3.9$s per query, which is faster than TFVTG ($4.7$s) and comparable to TAG ($3.4$s), while substantially outperforming both in accuracy (\cref{tab:iid}).
The efficiency gain comes from two sources: (1)~the DPP-based keyframe selection reduces the number of frames the LVLM must process, and (2)~adaptive routing directs simple queries to the Fast path, avoiding the multi-step reasoning overhead of TMP.
Compared to the Slow-only variant that applies TMP to every query, adaptive routing reduces inference time by $23.5\%$ ($5.1$s $\to$ $3.9$s) while also improving mIoU ($47.53$ $\to$ $48.93$, \cref{tab:ablation_component}), confirming that unnecessary reasoning on simple queries hurts both speed and accuracy.

\section{Conclusion}
\label{sec:conclusion}

We identified a reasoning gap in zero-shot video temporal grounding: feature-matching methods suffice for simple events but fail on complex queries requiring temporal and causal reasoning. We proposed \method{} to bridge this gap.
Built around a query-conditioned DPP, \method{} selects diverse, query-relevant keyframes and derives a spectral entropy-based difficulty indicator from a single decomposition, routing simple queries to direct prediction and complex queries to our Temporal Markup Prompting for structured reasoning.
Experiments on Charades-STA and ActivityNet Captions show that \method{} achieves state-of-the-art zero-shot performance under both IID and OOD settings, validating that deciding \emph{when} to reason is as important as \emph{how} to reason.

\bibliographystyle{splncs04}
\bibliography{references}

\end{document}